\documentclass[10pt,conference]{IEEEtran}

\usepackage[utf8]{inputenc}
\usepackage{graphicx}
\usepackage{amsmath}
\usepackage{booktabs} 
\usepackage{hyperref}
\usepackage{pifont}
\newcommand{\cmark}{\ding{51}}
\newcommand{\xmark}{\ding{55}}
\begin{document}
\title{Evaluating Financial Intelligence in Large Language Models: Benchmarking SuperInvesting AI with LLM Engines}


\author{
    \IEEEauthorblockN{
        Akshay Gulati\IEEEauthorrefmark{1}, 
        Kanha Singhania\IEEEauthorrefmark{1}, 
        Tushar Banga\IEEEauthorrefmark{1}, 
        Parth Arora\IEEEauthorrefmark{1}, 
        Anshul Verma\IEEEauthorrefmark{1}, \\
        Vaibhav Kumar Singh\IEEEauthorrefmark{1}, 
        Agyapal Digra\IEEEauthorrefmark{1}, 
        Jayant Singh Bisht\IEEEauthorrefmark{1}, 
        Danish Sharma\IEEEauthorrefmark{1}, 
        Varun\IEEEauthorrefmark{1},
        and Shubh Garg \IEEEauthorrefmark{1}
    }

    \IEEEauthorblockA{\IEEEauthorrefmark{1}The Future University}

    
    \thanks{Emails: \{akshay.gulati, kanha0, tushar, parth.arora, anshul.verma, vaibhav.dang, agyapal.digra, jayant.singh.bisht, danish.sharma, varun.singla, shubh.garg\}@thefuture.university}
}

\maketitle

\begin{abstract}
Large language models are increasingly used for financial analysis and investment research, yet systematic evaluation of their financial reasoning capabilities remains limited. In this work, we introduce the AI Financial Intelligence Benchmark (AFIB), a multi-dimensional evaluation framework designed to assess financial analysis capabilities across five dimensions: factual accuracy, analytical completeness, data recency, model consistency, and failure patterns. We evaluate five AI systems: GPT, Gemini, Perplexity, Claude, and SuperInvesting, using a dataset of 95+ structured financial analysis questions derived from real-world equity research tasks. The results reveal substantial differences in performance across models. Within this benchmark setting, \textit{SuperInvesting achieves the highest aggregate performance}, with an average factual accuracy score of 8.96/10 and the highest completeness score of 56.65/70, while also demonstrating the lowest hallucination rate among evaluated systems. Retrieval-oriented systems such as Perplexity perform strongly on data recency tasks due to live information access but exhibit weaker analytical synthesis and consistency. Overall, the results highlight that financial intelligence in large language models is inherently multi-dimensional, and systems that combine structured financial data access with analytical reasoning capabilities provide the most reliable performance for complex investment research workflows.

\end{abstract}

\section{Introduction}
Large language models (LLMs) have rapidly advanced the state of natural language understanding and reasoning. Recent models such as GPT-3\cite{brown2020language}, PaLM \cite{chowdhery2023palm},and nd GPT-4 \cite{achiam2023gpt}demonstrate strong performance across a wide range of language and reasoning tasks, enabling applications in domains including software engineering, scientific research, and professional decision support. Their capabilities are commonly evaluated using large-scale benchmark suites such as MMLU\cite{hendrycks2020measuring} and BIG-Bench\cite{srivastava2023beyond}, while broader evaluation frameworks such as HELM\cite{liang2022holistic} attempt to assess language models across multiple capability and risk dimensions. These evaluation efforts have provided valuable insights into general reasoning ability, knowledge coverage, and safety considerations in large-scale language models.

At the same time, recent work has highlighted important limitations in existing evaluation frameworks. Research on hallucination and factual reliability in language generation demonstrates that LLMs can produce fluent but factually incorrect outputs\cite{maynez-etal-2020-faithfulness,lin-etal-2022-truthfulqa,ji2023survey}.Benchmarks such as TruthfulQA\cite{lin-etal-2022-truthfulqa} attempt to quantify models’ tendency to generate misleading information, while interactive evaluation platforms such as Chatbot Arena\cite{zheng2023judging} measure model performance through human preference comparisons. However, these evaluations primarily focus on general knowledge or conversational tasks and do not capture the domain-specific requirements of high-stakes analytical workflows.

One domain where these limitations are particularly important is financial analysis. LLMs are increasingly used by investors, analysts, and financial institutions to retrieve financial metrics, interpret corporate disclosures, summarize macroeconomic developments, and generate investment theses. Financial reasoning differs from many standard NLP tasks in that it requires precise numerical accuracy, integration of multiple financial statements, awareness of recent market developments, and coherent analytical synthesis across heterogeneous data sources. Errors in such analyses can directly influence investment decisions, making reliable evaluation of AI systems in this domain especially important.

Recent research has begun exploring the application of language models in finance. Domain-adapted models such as FinBERT\cite{araci2019finbert} and BloombergGPT\cite{wu2023bloomberggpt} demonstrate that training language models on financial corpora can improve performance on financial NLP tasks such as sentiment analysis, document classification, and question answering. Similarly, emerging benchmarks such as FinanceBench\cite{wu2024bloomberggpt} evaluate the ability of LLMs to answer questions based on corporate financial disclosures. While these efforts represent important progress, most existing evaluations focus on isolated information retrieval or document understanding tasks, rather than the full analytical workflow required in professional investment research.

This gap motivates the need for evaluation frameworks that measure financial intelligence in AI systems more directly. In this paper we introduce the AI Financial Intelligence Benchmark (AFIB), a multi-dimensional evaluation framework designed to assess the capability of AI systems to perform financial analysis tasks relevant to capital markets. Rather than focusing on a single capability, AFIB evaluates models across five dimensions that reflect the practical requirements of financial research workflows: factual accuracy and hallucination resistance, analytical completeness, data recency and news integration, model consistency, and real-world failure patterns.

The benchmark evaluates five widely used AI systems: GPT, Gemini, Perplexity, Claude, and SuperInvesting, across 71 financial analysis queries covering corporate financial metrics, macroeconomic developments, and complex conglomerate analysis in the Indian equity market. In addition to controlled benchmark tasks, we incorporate a dataset of 432 negatively rated assistant responses collected from a production financial AI deployment, enabling the analysis of real-world failure modes that are rarely captured in conventional benchmarks.

Our empirical results reveal substantial capability differences across models. Retrieval-based systems demonstrate advantages in incorporating recent financial events, while domain-specialized models show stronger performance in numerical accuracy, analytical completeness, and cross-statement reasoning. These findings highlight that financial analysis performance in AI systems is multi-dimensional, and that architectures combining structured financial data access with analytical reasoning capabilities appear better suited for investment research tasks.

This work makes three primary contributions:
\begin{itemize}
    \item \textbf{A multi-dimensional benchmark for evaluating financial intelligence in AI systems}, capturing five capability dimensions relevant to professional investment analysis.
    \item \textbf{A curated evaluation dataset spanning 71 financial analysis queries and a deployment dataset of 432 real-world failure cases}, enabling both controlled benchmarking and empirical failure analysis.
    \item \textbf{A comparative evaluation of leading AI models}, identifying systematic strengths and limitations across accuracy, completeness, recency, consistency, and hallucination resistance.
\end{itemize}
Together, these results provide a structured framework for evaluating the reliability of AI systems used in financial analysis and highlight key challenges in deploying language models in high-stakes decision-making environments.
\section{Related Work}
The evaluation of large language models (LLMs) has become a central research problem as these systems are increasingly deployed in real-world decision-support settings. Prior work relevant to this study spans four main areas: general-purpose LLM benchmarking, evaluation frameworks and automated judging methods, reliability and hallucination analysis, and financial domain language models and benchmarks. While these lines of work have significantly advanced the evaluation of language models, existing approaches only partially capture the analytical requirements of professional financial research. While these lines of work have significantly advanced our understanding of LLM capabilities, they leave several aspects of financial analysis evaluation underexplored.
\begin{table*}[t]
\centering
\caption{Comparison of AFIB with existing financial LLM benchmarks.}
\label{tab:benchmark_comparison}

\begin{tabular}{l c c c c c c c}
\hline
\textbf{Benchmark} & \textbf{Year} & \textbf{FAT} & \textbf{NR} & \textbf{MDE} & \textbf{RFD} & \textbf{FA} & \textbf{RE} \\
\hline
FinQA\cite{chen2021finqa} & 2021 & \xmark & \cmark & \xmark & \cmark & \xmark & \xmark \\
FinanceBench \cite{islam2023financebench} & 2023 & \xmark & \cmark & \xmark & \cmark & \xmark & \xmark \\
FinBen  \cite{xie2024finben} & 2024 & \cmark & \cmark & \xmark & \cmark & \xmark & \xmark \\
FinanceQA \cite{mateega2025financeqa} & 2025 & \cmark & \cmark & \xmark & \cmark & \xmark & \xmark \\
FinAuditing \cite{wang2025finauditing} & 2025 & \cmark & \cmark & \xmark & \cmark & \xmark & \xmark \\
Fin-RATE \cite{jiang2026fin} & 2026 & \cmark & \cmark & \cmark & \cmark & \xmark & \xmark \\
\textbf{AFIB (Ours)} & \textbf{2026} & \cmark & \cmark & \cmark & \cmark & \cmark & \cmark \\
\hline
\end{tabular}

\vspace{0.5em}
\footnotesize{
\textbf{FAT}: Financial Analysis Tasks,
\textbf{NR}: Numerical Reasoning,
\textbf{MDE}: Multi-Dimensional Evaluation,
\textbf{RFD}: Real Financial Data,
\textbf{FA}: Failure Analysis,
\textbf{RE}: Recency Evaluation.
}

\end{table*}
\subsection{Benchmarking Large Language Models}
A large body of research has focused on developing benchmarks to measure the reasoning and knowledge capabilities of LLMs. One of the most widely used evaluation datasets is MMLU\cite{hendrycks2020measuring}, which measures model performance across 57 academic disciplines. Similarly, BIG-Bench\cite{srivastava2023beyond}aggregates hundreds of tasks designed to evaluate reasoning, commonsense knowledge, and linguistic competence.Several works have attempted to extend these benchmarks into more comprehensive evaluation frameworks. HELM\cite{liang2022holistic} introduced a holistic evaluation framework that assesses language models across multiple axes including accuracy, calibration, robustness, and fairness. Building on this idea, Chatbot Arena and MT-Bench \cite{zheng2023judging} evaluate conversational models using large-scale human preference comparisons, providing dynamic leaderboards for frontier models.More recently, several benchmarks have begun to explore domain-specific evaluation of LLM capabilities. 

FinBen \cite{xie2024finben} introduces a large-scale financial benchmark containing dozens of datasets spanning tasks such as financial information extraction, reasoning, and decision support. FinanceQA \cite{mateega2025financeqa} evaluates LLMs on complex financial analysis questions derived from investment research scenarios, demonstrating that many models struggle with multi-step financial reasoning. Other recent benchmarks such as FinAuditing \cite{wang2025finauditing} and Fin-RATE \cite{jiang2026fin} extend evaluation to structured financial documents and longitudinal financial analysis across multiple filings. These efforts highlight the growing interest in evaluating LLMs for financial applications, but typically focus on isolated tasks rather than the broader analytical workflows involved in investment research.

More recent research has explored automated and scalable evaluation approaches. AlpacaEval\cite{dubois2023alpacafarm} and its improved version AlpacaEval 2.0\cite{dubois2024length} propose automated evaluation frameworks calibrated against human judgments. Similarly,Arena-Hard\cite{li2024crowdsourced}introduces a challenging benchmark designed to stress-test frontier language models.Another emerging direction is the use of LLM-as-a-judge evaluation frameworks, where language models themselves are used to evaluate outputs of other models~\cite{chiang2024chatbot,zheng2023judging}.While these approaches significantly improve the scalability of evaluation, they remain primarily focused on general reasoning and conversational tasks rather than domain-specific analytical workflows.

\subsection{Reliability and Hallucination in LLM Systems}
The reliability of language model outputs has emerged as a major concern in recent research. Early work by Maynez et al. (2020) \cite{maynez-etal-2020-faithfulness}demonstrated that neural text generation systems frequently produce outputs that are fluent but factually incorrect. Building on this insight, TruthfulQA\cite{lin-etal-2022-truthfulqa}introduced a benchmark designed to measure whether language models reproduce common misconceptions.Subsequent work has explored systematic analysis of hallucination behavior in language models.Ji et al. (2023) \cite{ji2023survey} provide a comprehensive survey of hallucination phenomena in natural language generation. More recent work has introduced benchmarks specifically designed to evaluate hallucination in complex domains. PHANTOM \cite{jiphantom} evaluates hallucination detection in long-context financial question answering tasks, while FAITH \cite{zhang2025faith} analyzes hallucination behavior in tabular financial data extraction. These studies highlight the particular challenges associated with numerical reasoning and factual grounding in financial contexts.Additional studies have examined calibration and uncertainty in LLM outputs \cite{kadavath2022language} as well as emergent risks associated with large-scale language models \cite{ganguli2022predictability}.

More recent evaluation frameworks have focused on trustworthiness and safety. TrustLLM\cite{huang2024trustllm} proposes a multi-dimensional framework for evaluating trustworthiness in LLM systems, while Holistic LLM evaluation frameworks\cite{liang2022holistic,chiang2024chatbot} emphasize the importance of evaluating models across multiple reliability dimensions.However, these benchmarks typically evaluate hallucination and factuality in general knowledge tasks, rather than in contexts where numerical precision and temporal accuracy are critical, such as financial analysis.

\subsection{Financial NLP and Financial Language Models}
Parallel to developments in general LLM evaluation, researchers have explored the use of language models in financial applications. One of the earliest domain-adapted models is FinBERT\cite{araci2019finbert},which demonstrated improved performance on financial sentiment analysis tasks.More recently, large-scale financial language models have emerged. BloombergGPT\cite{wu2023bloomberggpt} is a domain-specific LLM trained on financial data and proprietary financial documents. Similarly, FinGPT~\cite{liu2023fingpt,yang2023fingpt} proposes an open-source framework for building financial AI systems that integrate LLMs with financial datasets and market information.

Financial reasoning datasets have also been proposed. FinQA\cite{chen2021finqa} evaluates models on numerical reasoning tasks involving financial tables, while ConvFinQA\cite{chen2022convfinqa} extends this dataset to conversational financial reasoning tasks. More recently, FinanceBench\cite{islam2023financebench} evaluates LLMs on question answering tasks derived from corporate financial reports.More recent benchmarks have explored financial decision-making and agent-based evaluation. InvestorBench \cite{li2025investorbench} evaluates LLM agents on investment strategy tasks such as portfolio construction and market trading simulations. Similarly, the Agent Market Arena benchmark \cite{qian2025agents} studies the behavior of LLM-based trading agents operating in simulated financial markets. 

These studies illustrate the growing interest in applying LLMs to financial decision-support systems.Recent surveys of LLM applications in finance\cite{nie2024survey,li2023large}highlight the growing use of language models for tasks such as earnings call analysis, automated financial reporting, and investment research support.Despite these advances, most existing financial NLP benchmarks focus on document understanding or question answering tasks, rather than the broader analytical workflows required in professional investment research.
\subsection{Gap in Existing Benchmarks}
Taken together, existing work reveals an important gap in the evaluation of AI systems for financial analysis. General-purpose benchmarks such as MMLU, BIG-Bench, and HELM evaluate broad reasoning capabilities but do not capture the domain-specific analytical processes required in professional financial research. Conversely, financial benchmarks such as FinQA, FinanceBench, FinBen, and FinanceQA primarily focus on document-level question answering or isolated reasoning tasks rather than evaluating the complete analytical workflow used by financial analysts.

In practice, investment research requires the integration of multiple financial statements, interpretation of financial metrics within business context, incorporation of recent market developments, and the construction of coherent analytical narratives.To address this gap, we introduce the AI Financial Intelligence Benchmark (AFIB), a framework designed to evaluate AI systems across multiple dimensions of financial intelligence, including numerical accuracy, analytical completeness, data recency awareness, reasoning consistency, and real-world failure patterns observed in deployed financial AI systems.Table~\ref{tab:benchmark_comparison} compares AFIB with recent benchmarks for financial language models.
\section{Methodology}
\begin{figure*}[t]
\centering
\includegraphics[width=\textwidth]{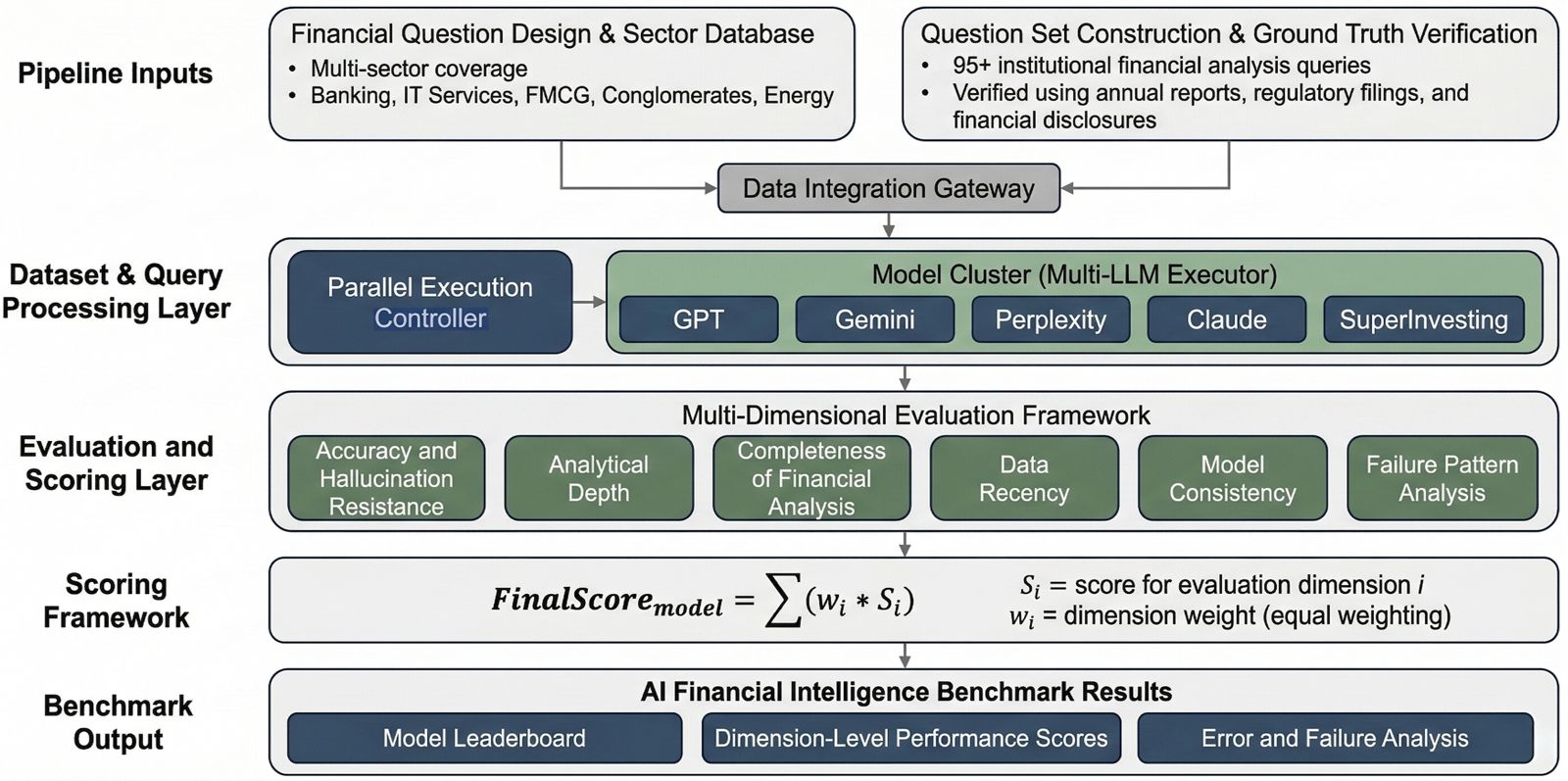}
\caption{Overview of the AI Financial Intelligence Benchmark (AFIB) evaluation pipeline. Financial analysis queries are constructed from multi-sector datasets and executed across multiple AI models. Model outputs are evaluated using a multi-dimensional scoring framework capturing accuracy, analytical depth, completeness, recency, consistency, and failure patterns.}
\label{fig:methodology}
\end{figure*}
\subsection{Benchmark Overview}
The AI Financial Intelligence Benchmark (AFIB) was developed to evaluate the capability of modern AI systems to perform financial analysis tasks comparable to those undertaken by professional investment analysts. Unlike conventional language model benchmarks that focus primarily on general reasoning or factual recall, financial analysis requires the integration of numerical accuracy, financial statement interpretation, macroeconomic awareness, and structured analytical reasoning. The benchmark therefore evaluates models across multiple complementary dimensions that together capture the analytical workflow of institutional equity research.Figure~\ref{fig:methodology} illustrates the overall AFIB evaluation pipeline, including the construction of financial query datasets, the parallel execution of queries across multiple AI models, and the multi-dimensional evaluation framework used to compute the final benchmark scores.

The evaluation framework was designed around five capability dimensions: factual accuracy and hallucination resistance, analytical depth and valuation logic, completeness of financial analysis, data recency and news integration, and consistency across repeated queries. Each dimension corresponds to a distinct analytical capability required in real-world financial research environments. These dimensions were implemented as independent benchmark modules that were applied consistently across all evaluated models.

The study evaluates five AI systems : GPT, Gemini, Perplexity, Claude, and SuperInvesting across structured financial analysis questions covering Indian listed equities. The dataset includes companies from multiple sectors of the Indian economy, including large-cap conglomerates, banking institutions, information technology services firms, consumer goods companies, and emerging industries such as renewable energy and defence manufacturing. This sectoral diversity ensures that models are tested across different business models, regulatory environments, and capital structures.
\subsection{Evaluation Dimensions}
The benchmark evaluates model performance across five capability dimensions representing the core analytical requirements of institutional financial research. Each dimension measures a distinct aspect of model behaviour, ranging from numerical reliability to the stability of reasoning across repeated queries. Together, these dimensions provide a comprehensive view of financial intelligence in AI systems.Rather than evaluating models on a single metric, the benchmark treats financial analysis as a multi-dimensional capability in which accurate data retrieval, analytical reasoning, contextual interpretation, and response stability must operate simultaneously. Each dimension therefore contributes to the overall evaluation of a model’s usefulness in financial research workflows.
\subsubsection{Factual Accuracy and Hallucination Resistance}
The first evaluation dimension measures the factual correctness of model outputs relative to verified financial ground truth. Financial analysis requires high numerical precision because incorrect figures relating to revenue, earnings, capital expenditure, or financial ratios can materially distort investment conclusions. In this context, hallucinations are defined as instances where a model generates a specific factual claim particularly , a numerical value that cannot be verified against authoritative financial sources.To evaluate accuracy, all numerical claims produced by the models were cross-checked against verified financial data from company filings and regulatory disclosures. Let $N$ denote the number of verifiable numerical claims produced by a model and $C$ denote the subset of those claims that match the ground truth. The factual accuracy score is therefore defined as:

\begin{equation}
\text{Accuracy} = \frac{C}{N}
\end{equation}

The hallucination rate represents the complement of this measure and captures the proportion of incorrect or fabricated numerical statements. In addition to numerical fabrication, hallucinations were categorized according to the nature of the error, including incorrect financial period attribution, misapplied metric definitions, and unsupported factual assertions.
Ground truth data was sourced exclusively from official financial documents including SEBI-regulated filings, stock exchange disclosures, and company annual reports for the relevant financial year.

\subsubsection{Analytical Depth and Valuation Logic}
Financial analysis requires the synthesis of multiple variables rather than the retrieval of isolated financial metrics. The analytical depth dimension therefore evaluates whether models demonstrate structured reasoning when interpreting financial data. Specifically, this dimension assesses whether models integrate interactions between financial variables, apply appropriate valuation frameworks, and demonstrate awareness of sector-specific dynamics.

Questions in this module required models to analyse financial relationships such as return on capital trends, margin compression, capital expenditure cycles, and valuation multiples. Responses were evaluated according to the depth of reasoning demonstrated in the analytical narrative. Models received higher scores when their responses reflected multi-variable reasoning linking financial metrics to business drivers, sector structure, and macroeconomic context.This evaluation distinguishes models that simply retrieve financial figures from those capable of constructing coherent investment reasoning comparable to that produced by human analysts.

\subsubsection{Completeness of Financial Analysis}
The completeness dimension evaluates whether model responses address all identifiable components of complex financial analysis questions. Many financial queries require the integration of multiple analytical elements, including financial metrics, contextual explanations, comparative analysis, and the interpretation of financial statements. A response that addresses only a subset of these elements may be directionally correct but remains analytically incomplete.To quantify completeness, responses were evaluated across a structured rubric measuring coverage of requested metrics, contextual interpretation, reasoning depth, narrative coherence, and evidence referencing. Let $k_q$ denote the number of analytical components required for question $q$, and let $k_{r,q}$ denote the number of components addressed in the model's response for that question. The completeness score for a single response is therefore defined as:

\begin{equation}
\text{Completeness}_q = \frac{k_{r,q}}{k_q}
\end{equation}

To obtain an overall completeness score across the benchmark dataset containing $Q$ questions, the per-question completeness scores are averaged:

\begin{equation}
\text{Completeness}_{model} = \frac{1}{Q}\sum_{q=1}^{Q} \text{Completeness}_q
\end{equation}

Questions in this module focused primarily on structurally complex companies such as Reliance Industries and ITC Ltd., whose diversified business segments require multi-statement analysis and segment-level reasoning. These companies therefore provide a particularly demanding test of analytical completeness.

\subsubsection{Data Recency and News Integration}
Financial markets are inherently forward-looking and highly sensitive to recent developments such as earnings announcements, regulatory changes, and macroeconomic policy decisions. Analytical outputs based on outdated information can therefore produce misleading conclusions even when the underlying reasoning is correct. The data recency dimension evaluates whether models incorporate current financial information into their analytical outputs.

The benchmark includes questions requiring knowledge of events occurring within the most recent financial reporting period, including quarterly earnings releases, Reserve Bank of India monetary policy decisions, sector-specific developments, and capital market activity. Responses were evaluated according to whether the model correctly identified the relevant financial period, incorporated recent events into its analysis, and interpreted their significance within the investment thesis.This dimension therefore captures the ability of AI systems to integrate real-time financial information into analytical reasoning rather than relying solely on static knowledge.

\subsubsection{Consistency Across Repeated Queries}
The final evaluation dimension measures the stability of model responses when identical or semantically equivalent questions are presented across multiple independent sessions. In financial research environments, analysts frequently revisit the same analytical queries over time, and reliable analytical tools must therefore produce stable outputs across repeated interactions.To evaluate response consistency, a subset of benchmark questions was submitted to each model across multiple independent sessions. Variations in numerical outputs, analytical reasoning structures, and final conclusions were recorded and analyzed.

Let $R_i$ denote the response generated during the $i$-th run of a repeated query, and let $V(R_i)$ represent the extracted numerical output or analytical verdict from that response.Consistency is measured by examining the variance of these outputs across repeated runs. Formally, the consistency score is defined as:

\begin{equation}
\text{Consistency} = 1 - \mathrm{Var}\big(V(R_i)\big)
\end{equation}

where $\mathrm{Var}(\cdot)$ denotes the statistical variance across all repeated responses for the same query. Higher consistency scores correspond to lower variance across repeated responses, indicating more stable and reproducible analytical behavior.

\subsection{Dataset Construction}
The benchmark dataset was constructed by domain specialists with experience in institutional equity research. Questions were designed to reflect the types of analytical tasks typically performed by professional investment analysts. The construction process followed three guiding principles: verifiability, representativeness, and analytical difficulty.Verifiability ensures that all questions involving financial metrics can be evaluated against authoritative financial data sources. Representativeness ensures that the dataset reflects realistic analytical tasks including financial metric retrieval, valuation reasoning, macroeconomic interpretation, and investment thesis development. Analytical difficulty ensures that the benchmark differentiates meaningfully between models with superficial retrieval capabilities and those capable of deeper financial reasoning.

The dataset spans multiple sectors of the Indian capital markets including banking, information technology services, consumer goods, conglomerates, and renewable energy companies. This diversity ensures that models are tested across different financial structures and industry dynamics.Ground truth data was obtained from publicly available financial documents including company annual reports, quarterly earnings releases, stock exchange disclosures, and macroeconomic publications issued by regulatory authorities.
\subsection{Evaluation Procedure}
All models were evaluated using a standardized evaluation protocol designed to ensure comparability across systems. Each benchmark question was submitted to every model using identical prompts without providing additional contextual information about the evaluation process. Responses were recorded in full and evaluated without modification.The evaluation process consisted of four stages. First, benchmark questions were submitted to each model under controlled conditions. Second, responses were extracted and stored for analysis. 

Third, numerical claims were verified against authoritative financial sources. Finally, responses were scored using structured scoring rubrics corresponding to the evaluation dimensions described earlier.For consistency testing, selected questions were repeated across multiple independent sessions without shared conversation history. This ensured that observed differences in model responses reflect inherent variability in model behaviour rather than contextual information from previous interactions.

\subsection{Scoring Framework}
The benchmark scoring framework aggregates performance across the five evaluation dimensions to produce a unified composite score. Each dimension produces a normalized score in the range $[0,100]$, enabling comparisons across different evaluation modules.Let $S_i$ denote the score of a model on evaluation dimension $i$, and let $w_i$ represent the weight assigned to that dimension. The overall benchmark score for a model is computed as a weighted aggregation:

\begin{equation}
\text{Score}_{\text{model}} = \sum_{i=1}^{5} w_i S_i
\end{equation}

In the present benchmark design, all five evaluation dimensions are assigned equal weights such that

\begin{equation}
w_i = 0.20 \quad \forall i \in \{1,2,3,4,5\}.
\end{equation}

This ensures that no single dimension dominates the overall evaluation outcome.Additional adjustments were applied in cases where models declined to answer a question or produced partially complete responses. Refusal to provide an answer resulted in zero scores for the relevant analytical dimensions while preserving hallucination resistance where appropriate. Partially complete responses were scored proportionally according to the fraction of requested analytical components addressed.
\subsection{Evaluation Setup}
All evaluations were conducted during the FY2025–26 financial reporting period. Models were accessed through their publicly available interfaces under default configuration settings to replicate the typical working environment of financial practitioners using AI tools for research.Financial ground truth data was obtained from authoritative sources including SEBI-regulated filings, stock exchange disclosures, company annual reports, and regulatory publications issued by the Reserve Bank of India and the Ministry of Finance. 

All financial values are reported in Indian Rupees unless otherwise specified.The evaluation framework was applied consistently across all benchmark modules to ensure comparability of results. Detailed scoring rubrics, benchmark questions, and raw model responses supporting the results of the study are available as supplementary materials.
\section{Results and Discussion}
\subsection{Overall Benchmark Performance}
\begin{figure}[t]
    \centering
    \includegraphics[width=0.95\linewidth]{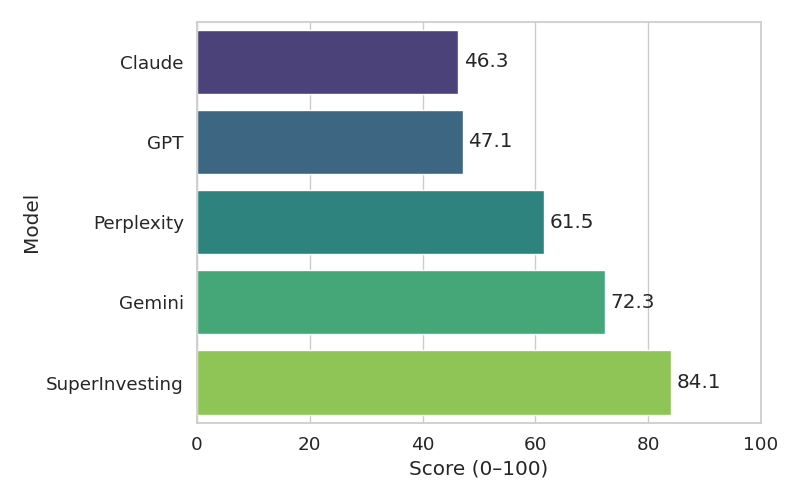}
    \caption{Overall Benchmark Leaderboard}
    \label{overall}
\end{figure}

The overall performance of the evaluated models is summarized in Figure \ref{overall}, which presents the composite benchmark scores aggregated across all evaluation dimensions. The results reveal clear separation between model capabilities, with SuperInvesting achieving the highest overall score, followed by Gemini and Perplexity. GPT and Claude rank lower due to limitations in accuracy, completeness, and real-time data access. The performance gap suggests that financial analysis tasks require a combination of numerical accuracy, structured reasoning, and contextual financial knowledge. Models optimized primarily for general reasoning or conversational capabilities appear less suited for institutional financial research workflows. The leaderboard therefore highlights the importance of domain specialization in financial AI systems.

\subsection{Dimension-Level Capability Analysis}
\begin{figure*}[t]
    \centering
    \includegraphics[width=0.95\linewidth]{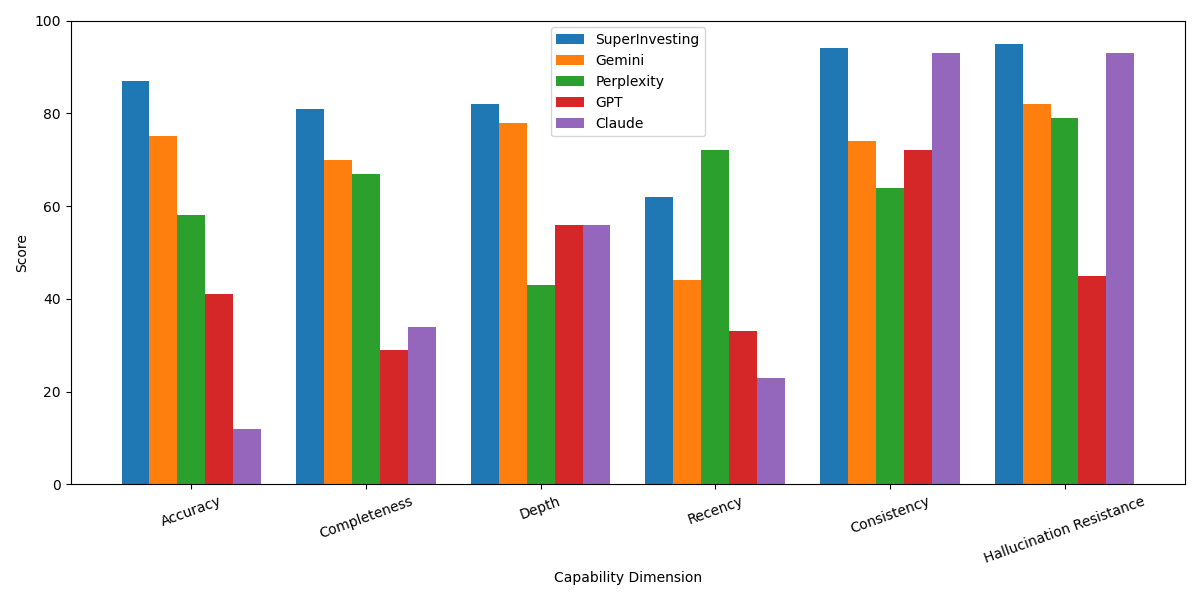}
    \caption{Model Capability Comparison Across Financial Benchmark Dimensions}
    \label{fig:dim}
\end{figure*}

To understand the sources of these performance differences, Figure \ref{fig:dim} presents a var graph comparison of model capability across the six evaluation dimensions used in the benchmark. SuperInvesting demonstrates consistently strong performance across most dimensions, particularly in accuracy, completeness, and analytical depth. Gemini shows competitive performance in reasoning-heavy tasks but exhibits weaker results in recency-dependent questions. Perplexity achieves the highest score in data recency but performs comparatively weaker in analytical reasoning and completeness. GPT and Claude display narrower capability profiles, with GPT demonstrating moderate reasoning ability but weaker accuracy, while Claude shows strong hallucination resistance but limited coverage across financial tasks. The radar comparison highlights that financial intelligence in AI systems is inherently multi-dimensional and cannot be captured by a single metric.
\subsection{Cross-Benchmark Leaderboard Comparison}
Table \ref{tab:benchmark_leaderboard} provides the per-benchmark ranking of models across the five evaluation modules used in the study. SuperInvesting ranks first in four out of five benchmarks, indicating consistent performance across independent evaluation frameworks. Gemini consistently ranks second or third, reflecting strong analytical reasoning capabilities among general-purpose models. Perplexity leads the data recency benchmark due to its live retrieval architecture but performs weaker in reasoning-intensive evaluations such as valuation depth. GPT and Claude frequently appear in lower ranks due to their respective limitations: hallucination frequency in GPT and systematic refusal behavior in Claude. These rankings confirm that the composite benchmark scores are not driven by isolated outcomes but instead reflect stable performance patterns across multiple evaluation methodologies.
\begin{table*}[ht]
\centering
\caption{Per-Benchmark Leaderboard Across Evaluation Dimensions}
\label{tab:benchmark_leaderboard}
\begin{tabular}{lccccc}
\hline
\textbf{Benchmark} & \textbf{1st} & \textbf{2nd} & \textbf{3rd} & \textbf{4th} & \textbf{5th} \\
\hline
Valuation Depth & SuperInvesting & Gemini & GPT & Claude & Perplexity \\
Consistency & SuperInvesting & Claude & Gemini & GPT & Perplexity \\
Accuracy / Hallucination & SuperInvesting & Gemini & Perplexity & GPT & Claude \\
Data Recency & Perplexity & SuperInvesting & Gemini & GPT & Claude \\
Completeness & SuperInvesting & Gemini & Perplexity & Claude & GPT \\
\hline
\end{tabular}
\end{table*}
\subsection{Trade-Off Between Recency and Analytical Depth}
\begin{figure}[t]
    \centering
    \includegraphics[width=0.95\linewidth]{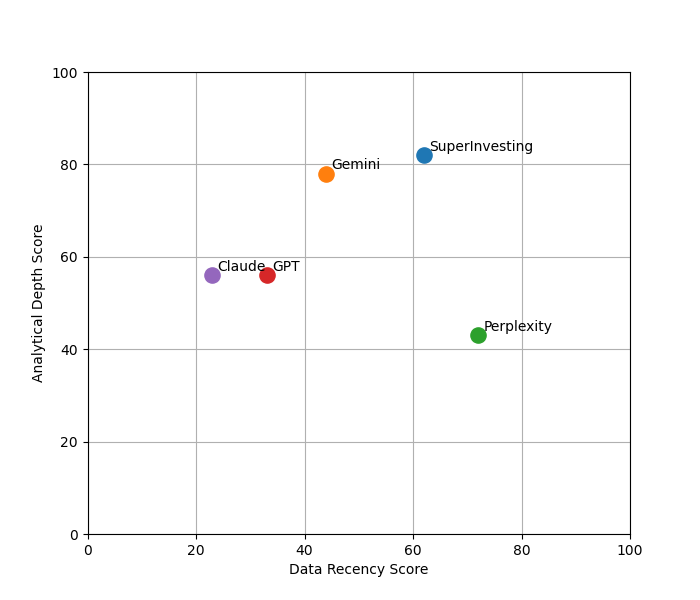}
    \caption{Trade-off Between Data Recency and Analytical Depth}
    \label{tradeoff}
\end{figure}

One of the most notable findings of the benchmark is the apparent trade-off between data recency capability and analytical reasoning depth. This relationship is illustrated in Figure \ref{tradeoff}, which plots model performance on the recency benchmark against analytical depth scores. Retrieval-oriented systems such as Perplexity demonstrate strong recency performance but comparatively weaker analytical reasoning. Conversely, reasoning-oriented models such as Gemini achieve high analytical depth but lack access to real-time financial information. SuperInvesting represents a partial exception to this pattern by achieving relatively strong performance on both dimensions. This observation suggests that hybrid architectures combining structured data pipelines with reasoning-oriented model design may provide a more effective approach for financial AI systems.
\subsection{Hallucination Risk in Financial Analysis}
\begin{figure}
    \centering
    \includegraphics[width=0.95\linewidth]{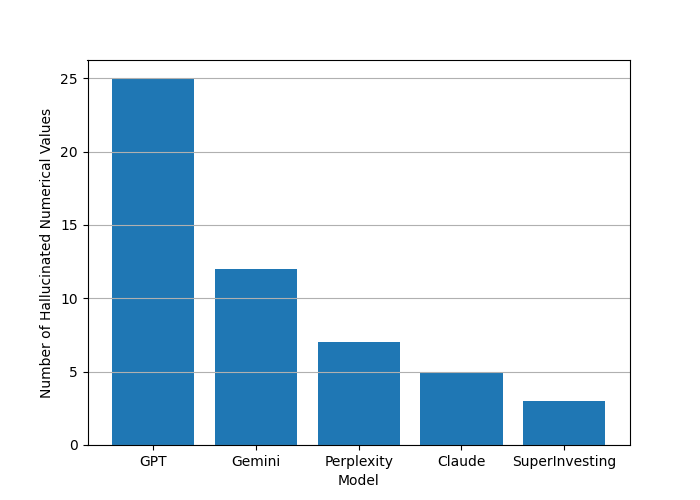}
    \caption{Hallucination Frequency Across Models}
    \label{hallucination}
\end{figure}

Hallucination behavior represents a critical reliability concern for financial AI systems. Figure \ref{hallucination} shows the frequency of hallucinated numerical financial values generated by each model during the accuracy benchmark evaluation. GPT exhibits the highest hallucination frequency, producing substantially more fabricated values than other models. Gemini and Perplexity show moderate hallucination rates, while Claude and SuperInvesting demonstrate stronger hallucination resistance. However, Claude achieves this primarily through conservative refusal strategies rather than improved numerical reasoning. From a financial risk perspective, confident numerical fabrication poses greater danger than refusal, as fabricated values may influence investment decisions if not independently verified.
\subsection{Multi-Dimensional Performance Patterns}
\begin{figure}
    \centering
    \includegraphics[width=1\linewidth]{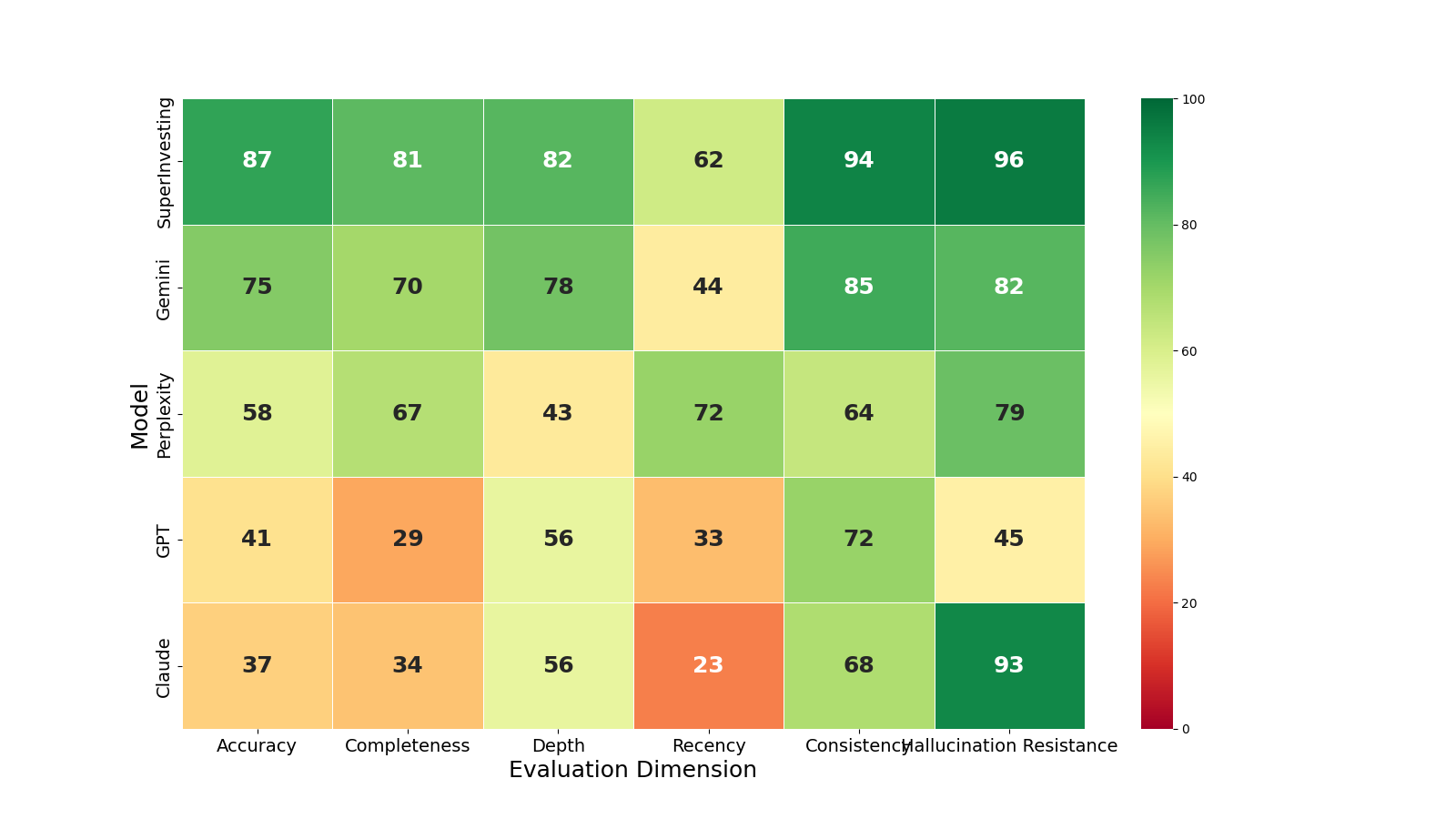}
    \caption{Heatmap of model performance across benchmark evaluation dimensions. Higher scores (green) indicate stronger performance.}
    \label{heatmap}
\end{figure}

The combined performance patterns across all evaluation dimensions are summarized in the heatmap shown in Figure \ref{heatmap}. This visualization provides a compact view of how each model performs across the benchmark’s six evaluation dimensions. SuperInvesting demonstrates consistently high scores across most dimensions, while Gemini exhibits strong analytical reasoning but weaker recency capability. Perplexity displays the opposite pattern, performing well in recency but weaker in reasoning and completeness. Claude achieves strong hallucination resistance but scores lower on completeness and recency, while GPT shows moderate reasoning ability but weak accuracy performance. The heatmap highlights that financial AI capability is not dominated by a single dimension but instead reflects the interaction between retrieval capability, reasoning depth, and numerical reliability.

\subsection{Implications for Financial AI Systems}
Taken together, the benchmark results suggest that no single architectural paradigm currently dominates all financial analysis tasks. Retrieval-based systems excel at integrating current financial information but may lack analytical synthesis capabilities. Reasoning-oriented models demonstrate stronger conceptual financial analysis but struggle with current-market data. Hybrid architectures that integrate structured financial data pipelines with domain-aware reasoning appear to produce the most reliable outputs for complex financial analysis workflows. These findings highlight the importance of specialized design in financial AI systems and suggest that domain-specific benchmarks are essential for evaluating AI performance in high-stakes decision-making environments.
\subsection{Robustness and Statistical Stability of Results}
To assess the robustness of the benchmark findings, we examined the stability of model performance across the five independent evaluation modules. The composite ranking remains largely consistent across benchmarks, as shown in Table \ref{tab:benchmark_leaderboard}, where SuperInvesting ranks first in four out of five evaluation categories. Gemini consistently occupies the second position in reasoning-oriented benchmarks, while Perplexity leads only the recency benchmark due to its live retrieval architecture. This consistency across independent evaluation tasks suggests that the final ranking is not driven by a single benchmark dataset but instead reflects stable capability differences between models.

In addition to cross-benchmark agreement, the distribution of scores across evaluation dimensions further confirms the stability of the results. As illustrated in the heatmap in Figure \ref{heatmap}, model performance patterns remain consistent across dimensions, with each model exhibiting a distinct capability profile. SuperInvesting demonstrates strong performance across most dimensions, Gemini excels in analytical reasoning, and Perplexity performs best in recency-based tasks. GPT and Claude show more limited capability profiles due to accuracy and coverage limitations respectively.

Finally, repeated-question experiments were conducted as part of the consistency benchmark to evaluate response stability. These tests revealed significant differences in output variance across models. SuperInvesting and Claude produced the most stable responses across repeated prompts, while Perplexity showed greater variability due to fluctuations in retrieved information. These observations reinforce the importance of consistency as a benchmark dimension for real-world financial workflows.

Taken together, the agreement across multiple benchmark modules, the stable dimension-level performance patterns shown in Figure \ref{fig:dim} and Figure \ref{heatmap}, and the response reproducibility results indicate that the benchmark findings are robust and not driven by isolated evaluation artifacts.Table~\ref{tab:appendix_model_responses} is showing some benchmakrking prompts and model responses in AIFB.
\begin{table*}[t]
\centering
\caption{Example benchmark prompts and model responses used in the AI Financial Intelligence Benchmark (AFIB). Responses are summarized for readability.}
\label{tab:appendix_model_responses}

\begin{tabular}{p{4cm} p{2.6cm} p{2.6cm} p{2.6cm} p{2.6cm} p{2.6cm}}
\hline
\textbf{Prompt} & \textbf{GPT} & \textbf{Gemini} & \textbf{Perplexity} & \textbf{Claude} & \textbf{SuperInvesting} \\
\hline

\textbf{Q1: Reliance EBITDA Contribution}

What percentage of Reliance’s consolidated EBITDA comes from Oil-to-Chemicals, Jio, and Retail in the latest financial year?

&
Provided segment percentages (43\%, 28\%, 29\%) but incorrectly concluded the three segments represent \textbf{100\%} of EBITDA.

&
Estimated segment contributions and explained structural shift toward consumer businesses. Combined estimate around \textbf{~88\%}.

&
Attempted calculation using annual report figures but produced inconsistent totals due to segment aggregation issues.

&
Provided historical estimates from earlier financial periods without latest-year verification.

&
Provided structured segment analysis with contextual explanation and estimated combined contribution around \textbf{~88–90\%}.

\\

\hline

\textbf{Q2: SBI vs ICICI NPA Comparison}

Compare the gross NPA and net NPA ratios of SBI and ICICI Bank for the most recent quarter and explain the role of provision coverage ratio (PCR).

&
Provided illustrative values and conceptual explanation of PCR but acknowledged figures were approximate.

&
Used recent financial results with structured explanation of GNPA, NNPA, and provisioning impact.

&
Retrieved figures from financial sources and presented a comparative table with PCR explanation.

&
Used older financial period data but correctly explained the PCR mechanism linking GNPA and NNPA.

&
Provided structured comparison of GNPA, NNPA, and PCR with explanation of asset quality differences and provisioning strategies.

\\

\hline

\textbf{Q3: L\&T Weighted Fundamental Score}

Construct a weighted fundamental score (out of 100) for Larsen \& Toubro using:

40\% order book-to-bill ratio  
30\% working capital days  
30\% EPC EBITDA margins.

Show exact calculation.

&
Assumed sector benchmarks and normalized metrics to compute a score of \textbf{60/100}. Used hypothetical values.

&
Applied financial benchmarks and calculated weighted score of \textbf{~83.7/100} with detailed reasoning.

&
Estimated financial inputs from disclosures and derived a score of \textbf{~79.3/100}.

&
Used sector averages and derived a score of \textbf{~82/100} with emphasis on working capital efficiency.

&
Calculated normalized benchmark scores and produced a weighted score of \textbf{~87.2/100} with detailed breakdown.

\\

\hline
\end{tabular}

\vspace{0.4em}
\footnotesize{Responses are summarized for brevity. Full model outputs used for scoring are available in the supplementary material.}

\end{table*}
\section{Limitations}
While the proposed benchmark provides a structured evaluation of AI systems for financial analysis, several limitations should be acknowledged. First, the evaluation is conducted within the context of Indian equity markets and focuses on a selected set of companies spanning sectors such as banking, information technology, consumer goods, and conglomerates. Although this sectoral diversity captures a wide range of financial structures, the findings may not generalize directly to other financial domains such as derivatives pricing, fixed income analysis, or global equity markets. Additionally, the benchmark primarily evaluates models through structured financial analysis questions rather than full end-to-end investment workflows. In practice, professional financial research often involves multi-step processes including document retrieval, spreadsheet modelling, and iterative hypothesis testing. Consequently, while the benchmark captures important components of financial reasoning and data interpretation, it does not fully replicate the complexity of institutional research environments.

A second limitation arises from the rapidly evolving nature of large language models and financial AI systems. The benchmark reflects model capabilities during a specific evaluation period and model versions, and performance characteristics may change as models receive architectural updates or expanded data integration capabilities. In particular, improvements in real-time retrieval systems and domain-specialized training may significantly alter model rankings in future evaluations. For this reason, the results should be interpreted as a snapshot of current capabilities rather than a definitive long-term ranking of financial AI systems.

\section{Future Work}
Several directions for future work emerge from the findings of this study. One important extension would be expanding the benchmark to cover a broader set of financial tasks, including portfolio construction, credit risk analysis, derivatives pricing, and macroeconomic forecasting. These domains require different forms of reasoning and quantitative modelling and would provide a more comprehensive assessment of financial intelligence in AI systems. Additionally, future benchmarks could incorporate longer-form financial documents such as earnings call transcripts, regulatory filings, and analyst research reports. Evaluating model performance on document-level reasoning tasks would better reflect real-world investment workflows where analysts must interpret large volumes of financial information.

Another promising direction involves exploring hybrid architectures that combine retrieval-based data access with structured financial reasoning capabilities. The results of this benchmark suggest that models optimized solely for retrieval or reasoning tend to exhibit complementary strengths and weaknesses. Systems capable of integrating real-time financial data pipelines with domain-aware reasoning frameworks may therefore achieve superior performance across multiple dimensions of financial analysis. Future research could investigate such hybrid approaches and develop benchmarks that evaluate the interaction between retrieval, reasoning, and numerical reliability.

\section{Conclusion}
This paper introduces the AI Financial Intelligence Benchmark, a multi-dimensional evaluation framework designed to assess the capability of AI systems to perform financial analysis tasks relevant to institutional investment research. By evaluating five widely used AI models across multiple analytical dimensions including analytical depth, factual accuracy, hallucination resistance, completeness, consistency, and data recency, the benchmark provides a systematic comparison of model performance in complex financial reasoning scenarios. The results demonstrate substantial capability differences across models and highlight that financial intelligence in AI systems cannot be adequately captured through single-metric evaluations.

More broadly, the findings reveal a structural trade-off between retrieval capability and analytical reasoning among current AI architectures. Retrieval-oriented systems excel at incorporating real-time financial information, while reasoning-oriented models demonstrate stronger performance in analytical interpretation and structured financial analysis. Hybrid systems that combine domain-specific data pipelines with financial reasoning frameworks appear to provide the most reliable performance across tasks. As AI systems continue to evolve, structured benchmarks such as the one presented in this study will be essential for understanding their strengths, limitations, and practical role in supporting financial decision-making workflows.
\bibliographystyle{unsrt}
\bibliography{library}
\appendix
\section{Example Benchmark Prompts and Model Responses}

This appendix presents representative prompts from the AI Financial Intelligence Benchmark (AFIB) along with responses generated by the evaluated models. These examples illustrate typical model behaviour observed during evaluation, including differences in numerical reasoning, analytical completeness, and factual reliability.

\subsection{Prompt 1: Reliance Industries Segment EBITDA Contribution}

\textbf{Prompt:}  
What percentage of Reliance’s consolidated EBITDA comes from Oil-to-Chemicals, Jio, and Retail in the latest financial year?

\begin{table*}[h]
\centering
\small
\begin{tabular}{p{3cm} p{11cm}}
\hline
\textbf{Model} & \textbf{Response (Excerpt)} \\
\hline

GPT &
Reported segment contributions as Oil-to-Chemicals 43\%, Jio 28\%, and Retail 29\%, and incorrectly concluded that these segments together account for \textbf{100\% of consolidated EBITDA}. \\

Gemini &
Provided a structured financial analysis estimating segment contributions around O2C $\sim$43\%, Jio $\sim$32\%, and Retail $\sim$13\%, concluding that these three segments contribute approximately \textbf{88\% of consolidated EBITDA}. \\

Perplexity &
Attempted a calculation using annual report figures but produced inconsistent totals due to incorrect aggregation of segment EBITDA values and reporting overlaps. \\

Claude &
Provided historical segment estimates from earlier financial periods and explained the role of segment diversification, but used outdated financial data. \\

SuperInvesting &
Produced a structured breakdown with contextual explanation of segment contributions (O2C $\sim$42–44\%, Jio $\sim$31–33\%, Retail $\sim$13–15\%), concluding that these segments contribute approximately \textbf{88–90\%} of consolidated EBITDA. \\

\hline
\end{tabular}
\caption{Example responses for Reliance Industries EBITDA composition prompt.}
\label{tab:appendix_reliance}
\end{table*}

\subsection{Prompt 2: SBI vs ICICI Bank Asset Quality Comparison}

\textbf{Prompt:}  
Compare the gross NPA and net NPA ratios of SBI and ICICI Bank for the most recent quarter. Explain how their provision coverage ratios (PCR) account for the difference.

\begin{table*}[h]
\centering
\small
\begin{tabular}{p{3cm} p{11cm}}
\hline
\textbf{Model} & \textbf{Response (Excerpt)} \\
\hline

GPT &
Provided approximate illustrative values for GNPA and NNPA and explained the concept of provision coverage ratio (PCR), but acknowledged that the figures were not based on the latest financial results. \\

Gemini &
Used recent financial data to compare asset quality metrics and explained the relationship between GNPA, NNPA, and PCR in determining bank balance sheet strength. \\

Perplexity &
Retrieved financial ratios from web sources and presented a comparison table of GNPA and NNPA values, linking the difference to provisioning behaviour. \\

Claude &
Used financial data from an earlier reporting period but provided a clear explanation of the relationship between PCR and the GNPA–NNPA gap. \\

SuperInvesting &
Produced a structured comparison including GNPA, NNPA, and PCR metrics, explaining how higher provisioning buffers reduce net NPA exposure and improve asset quality resilience. \\

\hline
\end{tabular}
\caption{Example responses for SBI vs ICICI Bank asset quality analysis prompt.}
\label{tab:appendix_banking}
\end{table*}

\subsection{Prompt 3: Asian Paints ROIC vs WACC Analysis}

\textbf{Prompt:}  
Asian Paints has historically commanded a high PE premium. Analyze whether this premium is justified by comparing its Return on Invested Capital (ROIC) and Weighted Average Cost of Capital (WACC) over the last five years.

\begin{table*}[h]
\centering
\small
\begin{tabular}{p{3cm} p{11cm}}
\hline
\textbf{Model} & \textbf{Response (Excerpt)} \\
\hline

GPT &
Outlined the methodology for computing ROIC and WACC and provided illustrative estimates showing ROIC consistently exceeding WACC by roughly 9–11 percentage points. \\

Gemini &
Declined to provide a detailed calculation due to insufficient data availability and instead discussed valuation context using ROE and ROCE indicators. \\

Perplexity &
Presented a detailed financial analysis including approximate historical ROIC and WACC values and discussed the implications of a narrowing value spread. \\

Claude &
Estimated historical ROIC and WACC values and concluded that Asian Paints maintains a significant positive ROIC–WACC spread supporting its premium valuation. \\

SuperInvesting &
Produced a structured capital efficiency analysis showing ROIC significantly exceeding WACC but highlighted a declining spread due to increasing competition and margin pressure. \\

\hline
\end{tabular}
\caption{Example responses for Asian Paints ROIC–WACC valuation analysis prompt.}
\label{tab:appendix_asianpaints}
\end{table*}
\end{document}